\documentclass[10pt,twocolumn,letterpaper]{article}

\usepackage[accsupp]{axessibility}  % Improves PDF readability for those with disabilities.
\usepackage[table,xcdraw]{xcolor}
\usepackage{cvpr}              % To produce the CAMERA-READY version

\usepackage[most,skins,theorems]{tcolorbox}
\tcbset{
  aibox/.style={
    width=\linewidth,
    top=8pt,
    bottom=4pt,
    colback=blue!6!white,
    colframe=black,
    colbacktitle=black,
    enhanced,
    center,
    attach boxed title to top left={yshift=-0.1in,xshift=0.15in},
    boxed title style={boxrule=0pt,colframe=white,},
  }
}
\newtcolorbox{AIbox}[2][]{aibox,title=#2,#1}

\definecolor{cvprblue}{rgb}{0.21,0.49,0.74}
\usepackage[pagebackref,breaklinks,colorlinks,allcolors=cvprblue]{hyperref}
\usepackage{multirow}

\def\confName{CVPR}
\def\confYear{2026}

\title{SpatialReward: Verifiable Spatial Reward Modeling for Fine-Grained Spatial Consistency in Text-to-Image Generation}

\author{
   \vspace{-0.8cm} \\
    Sashuai Zhou$^{1,2}$\thanks{Equal contribution.} ,
    Qiang Zhou$^{2}$\footnotemark[1],
    Junpeng Ma$^{3}$\footnotemark[1], 
    Yue Cao$^{2}$, Ruofan Hu$^{1}$, Ziang Zhang$^{1}$, \\ 
    Xiaoda Yang$^{1}$, Zhibin Wang$^{2}$, 
    Jun Song$^{2}$\thanks{Corresponding author.},
    Cheng Yu$^{2}$, Bo Zheng$^{2}$, 
    Zhou Zhao$^{1}$\footnotemark[2] \\
    \\
   \vspace{-0.8cm} \\
   {$^1$}Zhejiang University ,
   {$^2$}Alibaba Group,
   {$^3$}Fudan University\\
}

\begin{document}
\maketitle
\begin{abstract}
% Text-to-Image (T2I) generation has advanced rapidly, yet aligning outputs with complex, detailed prompts remains challenging. Current reward models for reinforcement learning or preference alignment often rely on fixed prompt formats and coarse global similarity scores, limiting generalization and overlooking fine-grained, verifiable details—leading to reward hacking, where images satisfy only partial constraints. We present a unified fine-grained reward framework that uses a vision-language model to decompose free-form prompts into structured representations encompassing objects, attributes, spatial relations, styles, and text. Guided by this decomposition, specialized verifiers—including object/attribute detectors, spatial-relation checkers, OCR, and image quality assessors—generate interpretable sub-scores, which are adaptively fused into an objective final reward. Experiments on diverse prompts and generative models (e.g., Stable Diffusion 3.5, Flux) demonstrate superior generalization, multi-constraint fidelity, and robustness to reward hacking over CLIPScore and direct VLM scoring. By grounding evaluation in explicitly verifiable visual evidence, our approach offers a transparent and reliable reward mechanism for T2I and broader cross-modal generation.
Recent advances in text-to-image (T2I) generation via reinforcement learning (RL) have benefited from reward models that assess semantic alignment and visual quality. However, most existing reward models pay limited attention to fine-grained spatial relationships, often producing images that appear plausible overall yet contain inaccuracies in object positioning. In this work, we present \textbf{SpatialReward}, a verifiable reward model explicitly designed to evaluate spatial layouts in generated images. SpatialReward adopts a multi-stage pipeline: a \emph{Prompt Decomposer} extracts entities, attributes, and spatial metadata from free-form prompts; expert detectors provide accurate visual grounding of object positions and attributes; and a vision-language model applies chain-of-thought reasoning over grounded observations to assess complex spatial relations that are challenging for rule-based methods. To more comprehensively evaluate spatial relationships in generated images, we introduce \textbf{SpatRelBench}, a benchmark covering object attributes, orientation, inter-object relations, and rendered text placement. Experiments on Stable Diffusion and FLUX show that incorporating SpatialReward into RL training consistently improves spatial consistency and overall generation quality, with results aligned more closely to human judgments. These findings indicate that verifiable reward models hold considerable potential for enabling more accurate and controllable optimization in text-to-image generation models. The project page is available
at: \url{https://github.com/LivingFutureLab/SpatialReward}

\end{abstract}

\section{Introduction}
\label{sec:intro}

Recent advances in text-to-image generation~\cite{Zero-shot-text-to-image-generation,latent-diffusion,flow-matching,caorepldm,Scaling-rectified-flow-transformers,wang2026geodesicnvs} have been increasingly fueled by reinforcement learning techniques~\cite{flowgrpo,reinforcementfine-tuningt2i,xue2025dancegrpo,li2025mixgrpo,wallace2024diffusion,Reward-ranked-finetuning,zhu2026medeyes}. Among these, GRPO-based approaches~\cite{flowgrpo,xue2025dancegrpo,li2025mixgrpo} have demonstrated notable effectiveness in improving the generative performance. A central component in these methods is the pre-trained reward model (RM)~\cite{imagereward,unifiedreward,hpsv2,pickscore,ghosh2023geneval}, which assesses both visual quality and semantic alignment, providing crucial feedback for policy-gradient optimization. 
Such reward-driven training has led to images that better match human preferences in global appearance and content.

Despite these advances, existing RMs primarily focus on global semantics and coarse visual quality, while neglecting fine-grained spatial relationships. As a result, current T2I models still struggle to preserve spatial consistency among objects within a scene~\cite{Benchmarking-spatial,T2i-compbench++,understanding-hallucinations-in-diffusion,Enhancing-prompt-understanding,Improving-spatial-consistency-int2i,Make-it-count}. These spatial inconsistencies degrade the realism of generated images and compromise faithful adherence to prompt semantics.

\begin{figure}[!t]
  \centering
  \includegraphics[width=0.98\columnwidth]{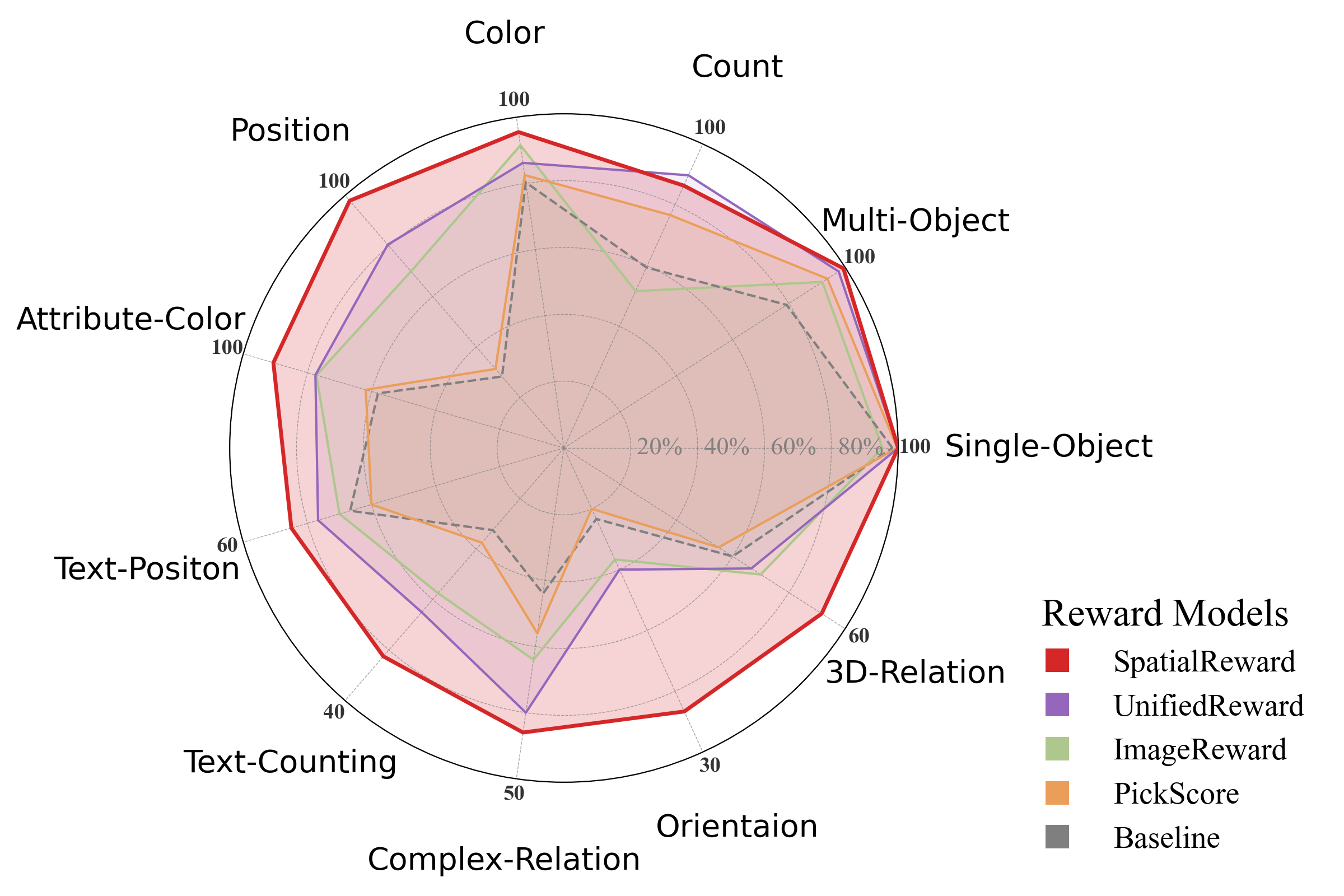}

\caption{Performance comparison of SD3.5-M~\cite{sd3.5} optimized via RL using SpatialReward versus Baseline Rewards.}
\label{fig:demo}
  \vspace{-1mm} % 
\end{figure}

\begin{figure*}[!t]
  \centering
  % NOTE: Remember to replace 'placeholder_figure.pdf' with your actual figure file.
  \includegraphics[width=\linewidth]{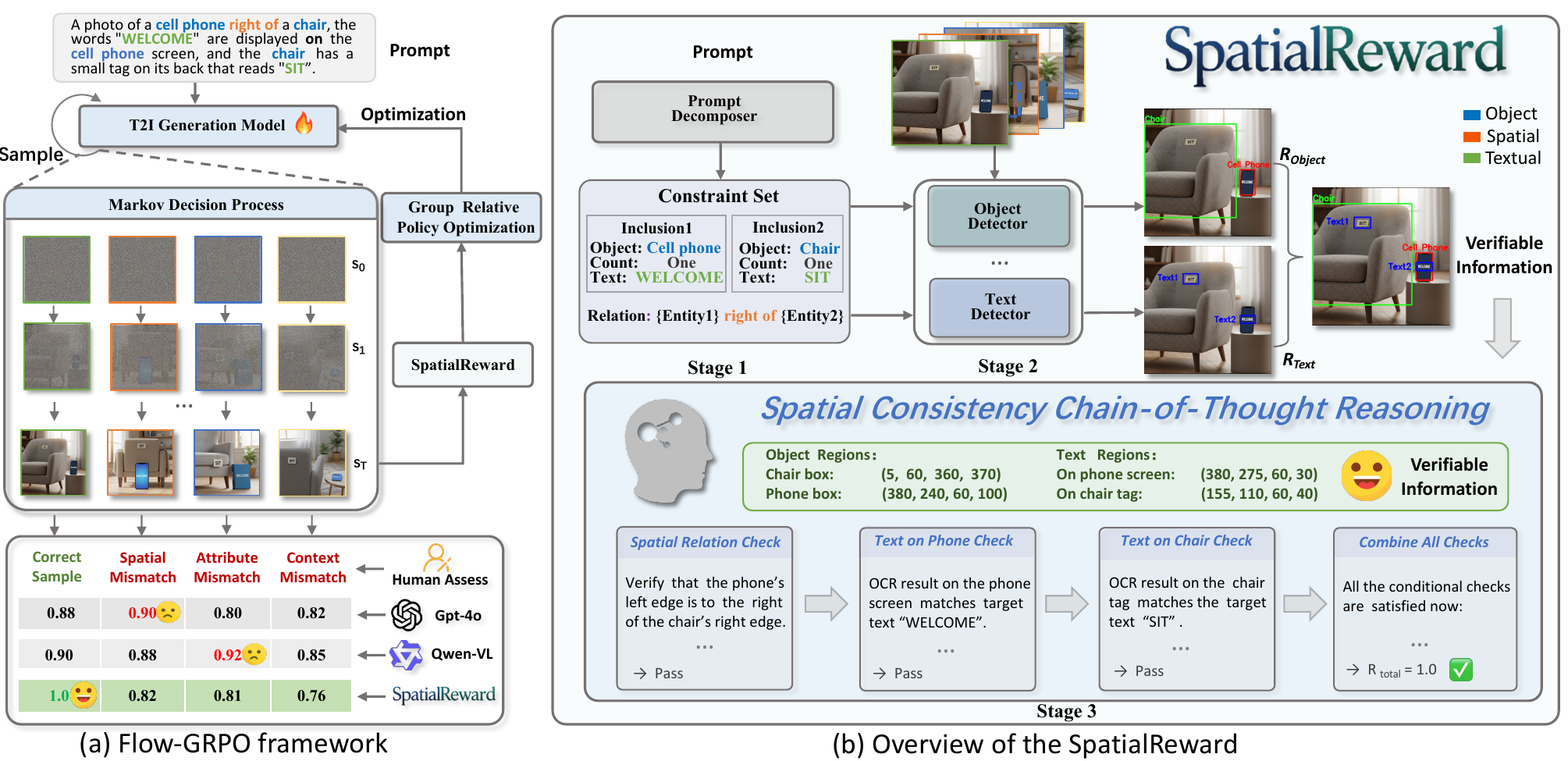} 
  \caption{
    \textbf{Overall framework of our approach.} 
    (a) Standard Flow-GRPO~\cite{flowgrpo} reinforcement learning pipeline for text-to-image generation. 
    (b) The proposed SpatialReward, which parses prompts into structured spatial and attribute constraints, verifies them on generated images via expert detection, and uses vision–language chain‑of‑thought reasoning to produce the final reward score.
    }

  \label{fig:framework}
\end{figure*}

We hypothesize that further improvements in T2I spatial generation depend more on verifiable, spatially-aware reward models than on refinements to RL training strategies. In particular, existing approaches suffer from two major deficiencies: 
\emph{Prompt-side rigidity}\textemdash Structured evaluation methods~\cite{ghosh2023geneval,t2i-compbench,T2i-compbench++}, such as Geneval~\cite{ghosh2023geneval}, rely on fixed-format prompts and predefined object detectors. For instance, they can handle template-based inputs like “a photo of a purple backpack” but fail to generalize to complex, compositional prompts frequently found in open-ended generation tasks. \emph{Vision-side overlooking}\textemdash Holistic evaluation methods based on CLIPScore~\cite{clip,pickscore,hpsv2,imagereward} or vision–language models~\cite{qwen-vl,unifiedreward,unifiedreward-think,xu2024visionreward} can handle arbitrary prompts and capture global semantics, but without fine-grained spatial verification they fail to detect positional errors, often rewarding scenes that are visually plausible yet spatially wrong~\cite{wang2023equivariant,hallucination,lin2023revisiting,ma2023crepe,yuksekgonul2022and}.

In this paper, we propose \textbf{SpatialReward}, a verifiable reward model explicitly built for fine-grained evaluation of spatial layouts.  
\emph{Regarding the prompt side}, we introduce a Prompt Decomposer to extract core entities, attributes, and spatial relations from arbitrary free-form prompts. This normalization enables expert detectors to operate regardless of prompt format, thereby enriching the model’s textual perception and supporting performance improvements in diverse generation scenarios~\cite{{chen2024spatialvlm,tang2019learning,Enhancing-semantic-fidelity,Region-controlled-text}}. 
\emph{Regarding the visual side}, we draw inspiration from the success of rule-based, verifiable rewards in logical reasoning~\cite{guo2025deepseekr1,yang2025r1,liu2025visualr1}, where explicit and checkable rewards have been shown to significantly improve complex inference performance. We extend this concept to visual spatial evaluation with a collaborative verification mechanism. First, object-related metadata extracted by the Prompt Decomposer is passed to open-set detectors~\cite{liu2024groundingdino,cui2025paddleocr30technicalreport,chung2024depth,li2022ppocr}, which produce factual and highly verifiable information on object attributes and locations, thereby reducing hallucinations. Since spatial relationship assessment requires a sequence of reasoning steps that link detected facts to relative positions and overall layouts, we incorporate the verified grounding into a chain-of-thought (CoT) process~\cite{cot0,cot1,vr-think} within a vision–language model. This explicit reasoning enables robust reward estimation for complex layouts and achieves greater flexibility than conventional rule-based checks.

To enrich the evaluation dimensions for spatial consistency in T2I models, we introduce \textbf{SpatRelBench}. This benchmark extends assessment beyond simple positional or color attributes to include object orientation, multi-object 3D positioning, complex spatial arrangements, and the placement of rendered text. We integrate our SpatialReward model into the Flow-GRPO~\cite{flowgrpo} framework, an RL approach built upon GRPO, using Stable Diffusion~\cite{sd3.5} and FLUX~\cite{flux} as base models. Experimental results show that our approach significantly enhances the spatial consistency of generated images.

Our main contributions are as follows:
\begin{itemize}
    \item We present \textbf{SpatialReward}, a verifiable spatial reward model, combining prompt decomposition, expert detection, and chain-of-thought reasoning.
    \item We introduce \textbf{SpatRelBench}, a benchmark extending spatial evaluation to fine-grained object attributes, inter-object relations, and rendered-text placement.    
    \item We conduct extensive experiments to verify that reinforcement learning with a verifiable reward model can significantly enhance spatial consistency in T2I generation.

\end{itemize}

\section{Related Work}
% \subsection{RL-based T2I Optimization}
% Diffusion-based T2I models, such as Stable Diffusion~\cite{sd3.5} and FLUX~\cite{flux}, have achieved high visual quality, while reinforcement learning has been shown to effectively improve generation performance across diverse prompts. Typical approaches train preference-based reward models to guide generation~\cite{ddpo,refl}, and recent work has extended this framework to GRPO-based method~\cite{flowgrpo,ye2025schedule,xue2025dancegrpo,guo2025deepseekr1}, thereby enhancing the generalization capacity of text-to-image generation. The potential of reinforcement learning motivates our design of a verifiable reward model focused on fine-grained spatial relationships in generated images.

\subsection{RL-based T2I Optimization}
Diffusion-based T2I models, such as Stable Diffusion~\cite{sd3.5} and FLUX~\cite{flux}, have achieved high visual quality, while reinforcement learning has been shown to effectively improve generation performance across diverse prompts. Typical approaches train preference-based reward models to guide generation~\cite{ddpo,refl}, and recent work has further adapted GRPO-style optimization to text-to-image generation~\cite{flowgrpo,ye2025schedule,xue2025dancegrpo}, improving text-image alignment and generalization. These advances are broadly relevant to multimodal applications that require fine-grained understanding, semantic alignment, and knowledge-enhanced reasoning~\cite{zhu2025pathology,Huang_2025_ICCV,wang2025ladb,yang2025omnicam,huang-etal-2025-enhancing-multimodal}. This trend motivates our design of a verifiable reward model specialized for fine-grained spatial relationship assessment in generated images.

\subsection{Reward Models for T2I Generation}
Existing reward models can be broadly classified into structured methods~\cite{ghosh2023geneval,t2i-compbench,T2i-compbench++} and holistic scorer methods~\cite{clip,imagereward,unifiedreward,hpsv2,pickscore,xu2024visionreward}. Structured approaches, such as Geneval~\cite{ghosh2023geneval}, rely on fixed-format prompts and predefined object detectors to verify attributes and spatial arrangements, achieving high precision within narrow templates but generalizing poorly to free-form inputs. Holistic scorers include CLIP-based regressors such as ImageReward~\cite{imagereward}, PickScore~\cite{pickscore}, and HPSv2~\cite{hpsv2}, which fine-tune CLIP to predict human-preference scores. More recently, vision-language model (VLM) backbones~\cite{qwen-vl,li2024llava} have been adopted to capture richer semantic and spatial cues, with representative methods including VisionReward~\cite{xu2024visionreward} and UnifiedReward~\cite{unifiedreward,Unifiedcot}. While VLM-based RMs improve flexibility over structured pipelines, they often overlook detailed spatial inconsistencies, highlighting the importance of verifiable reward models with robust spatial reasoning to ensure trustworthy evaluation across diverse prompts.

\subsection{Benchmarks for T2I Evaluation}
Early evaluation of image generation models relied on generic metrics such as Fréchet Inception Distance (FID)~\cite{FID}, Inception Score (IS)~\cite{IS}, and CLIPScore~\cite{clip,hpsv2,pickscore}, which, while effective for assessing overall image quality, fall short in capturing fine-grained image–text alignment and complex spatial relations. To remedy these issues, specialised benchmarks have emerged. GenEval~\cite{ghosh2023geneval} and T2I-CompBench~\cite{t2i-compbench,T2i-compbench++} are object-centric benchmarks focusing on fundamental aspects such as object attributes and positions, typically using object detectors for automated scoring. More recent benchmarks~\cite{meng2024phybench,fu2024commonsense,niu2025wise} employ advanced VLMs~\cite{xu2024visionreward,qwen-vl} to evaluate specific reasoning skills. However, current spatial benchmarks tend to ignore fine-grained inter-object relations, including orientation, 3D spatial positioning, and text placement. These gaps motivated us to develop SpatRelBench, a benchmark with richer spatial relation coverage.

\begin{figure*}[t]
  \centering
  \includegraphics[width=\textwidth]{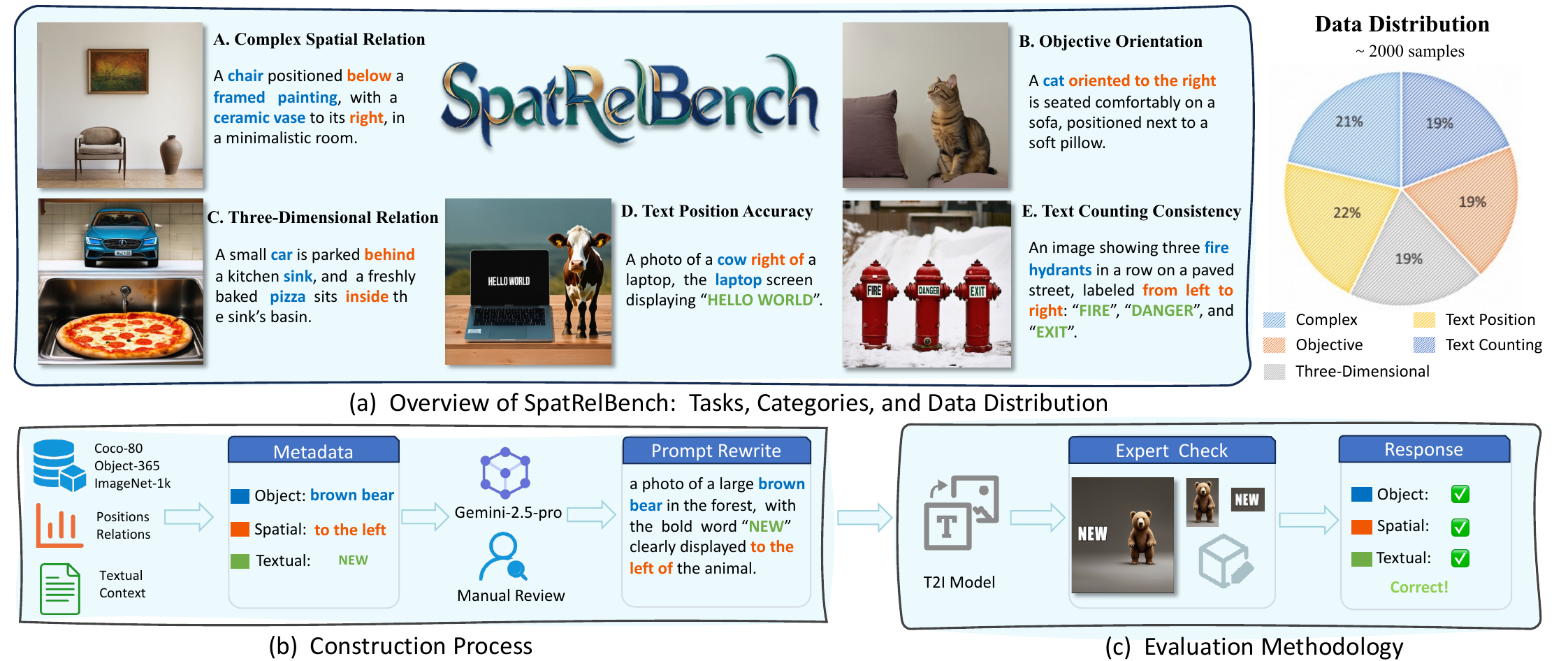}
  \caption{Overview of SpatRelBench, depicting benchmark tasks and their data distribution (a), the construction pipeline (b), and the evaluation methodology (c) designed to assess spatial relation understanding in text‑to‑image models.}
  \label{fig:benchmark}
\end{figure*}

% \section{SpatialReward}
% \label{sec:methodology}

% Our reward model methodology is designed to overcome the limitations of holistic evaluation by decomposing the reward calculation into three distinct, sequential stages. First, we parse the natural language prompt $P$ into a structured set of constraints $\mathcal{C}$. Second, we use specialized, high-precision models to verify each atomic constraint against the generated image $I$, yielding a vector of sub-rewards. Finally, we employ a Vision-Language Model (VLM) for advanced spatial reasoning and aggregate all signals into a final reward score $R(P, I)$. We detail each of these components below.

\section{SpatialReward}
\label{sec:methodology}

% SpatialReward decomposes reward computation into three sequential stages: 
% Section~\ref{sec:decomposition}, we transform the natural language prompt $P$ into a structured representation of spatial and attribute constraints $\mathcal{C}$. In Section~\ref{sec:extraction}, we employ specialized detection models to verify each constraint against the generated image $I$, producing fine-grained, verifiable signals. In Section~\ref{sec:reasoning}, a vision-language model leverages these signals through chain-of-thought reasoning to perform advanced spatial inference and aggregate them into the final reward score $R(P, I)$.

SpatialReward operates in three stages: Section~\ref{sec:decomposition} parses the prompt into structured spatial and attribute constraints; Section~\ref{sec:extraction} verifies these constraints on the generated image using expert detection models, yielding verified rewards; Section~\ref{sec:cot_reasoning} leverages a vision–language model with chain‑of‑thought reasoning to infer spatial relations and aggregate the results into the final reward score.

\subsection{Prompt Decomposition}
\label{sec:decomposition}

For RL-based training to generalize effectively, it is essential to handle prompts describing complex spatial arrangements of multiple objects~\cite{Benchmarking-spatial,t2i-compbench,ghosh2023geneval}. 
The spatial reward in our framework aims to evaluate such cases by extracting accurate inter-object positional relations for subsequent verification using expert detection models. 
However, reliable detection requires that the prompt content be first decomposed into structured elements: separating each subject, its attributes, and the spatial relations involved.

Free‑form prompts in the wild often contain irrelevant context or merge descriptions of different objects, introducing ambiguity that undermines detection accuracy. To mitigate this, we introduce a \textbf{Prompt Decomposer} $\mathcal{D}$ that transforms a free-form prompt $P$ into structured constraint set $\mathcal{C}$:
\begin{equation}
    \mathcal{C} = \mathcal{D}(P) = (\text{tag}, \mathcal{C}_{\text{inc}}, \mathcal{C}_{\text{exc}}),
\end{equation}
where \textit{tag} denotes the primary evaluation category (e.g., counting, orientation, spatial relation), $\mathcal{C}_{\text{inc}}$ contains inclusion constraints, and $\mathcal{C}_{\text{exc}}$ contains exclusion constraints. 
Each atomic constraint specifies object category, quantity, attributes, spatial relations, or textual inscriptions.

Inspired by metadata-driven evaluation frameworks~\cite{t2i-compbench,ghosh2023geneval}, we constructed a dataset of approximately 100k multi-object metadata instances, explicitly defining attributes, counts, spatial relations, and associated text for each subject. Using GPT-4o~\cite{openai2023gpt4}, we generated diverse natural language prompts from these metadatas, yielding (prompt, metadata) pairs for supervised training. We fine-tuned a Qwen2.5-VL-7B~\cite{qwen-vl} model to accurately extract core meta-attributes from unrestricted prompts. This structured decomposition provides a reliable foundation for subsequent fine-grained image evaluation in our pipeline.

\subsection{Fine-grained Verifiable Rewards}
\label{sec:extraction}
While the capabilities of VLMs are advancing rapidly, existing work~\cite{kamath2023text,lin2023revisiting,ma2023crepe,wang2023equivariant,yuksekgonul2022and,wang2025spatialclip} has shown that even state-of-the-art models struggle with compositional text prompts involving multiple objects, attribute binding, spatial-action relationships, counting, and logical reasoning. Relying solely on VLMs for these tasks makes it difficult to obtain stable and verifiable objective reward scores. Fortunately, modern open-domain object detection~\cite{liu2024groundingdino,ren2024groundedsam,li2022grounded,cheng2024yolo-grounding,wang2024orientanythinglearningrobust} and Optical Character Recognition (OCR) models~\cite{li2022ppocr,cui2025paddleocr30technicalreport,wei2024generalocr} demonstrate accuracy that significantly surpasses the judgmental capabilities of VLMs, providing objective scores that closely align with human evaluation standards. 
Leveraging the constraints extracted by the Prompt Decomposer in Section~\ref{sec:decomposition}, we integrate these specialized detectors to perform precise, criterion‑specific verification. This process yields a sub‑reward for each positive constraint 
$c_i \in \mathcal{C}_{\text{inc}}$, providing a verifiable and quantitatively accurate signal for assessing spatial positions, object attributes, and other fine‑grained relationships.

\paragraph{Object Attribute and Presence Reward.}
For each inclusion constraint $c \in \mathcal{C}_{\text{inc}}$ that refers to an object, we evaluate whether the generated image $I$ satisfies the specified visual and spatial properties, including object category, color, target count, orientation, and depth ordering. These properties are extracted via specialized detection models~\cite{liu2024groundingdino,cheng2024yolo-grounding,wang2024orientanythinglearningrobust,yang2024depth}.  

Given a target category in $c$, object detector $\mathcal{F}_{\text{det}}$~\cite{liu2024groundingdino,cheng2024yolo-grounding} is applied to the generated image $I$, yielding candidate bounding boxes 
\begin{equation}
    D_c = \{(B_j, s_j)\}_{j=1}^k,
\end{equation}
where $B_j$ denotes a bounding box and $s_j \in [0,1]$ its confidence score.  
Applying a confidence threshold $\tau_{\text{det}}$ produces the verified set of detections
\begin{equation}
    \mathcal{B}_c = \{ B_j \mid (B_j, s_j) \in D_c \ \land \ s_j \geq \tau_{\text{det}} \},
\end{equation}
whose cardinality $\hat{N}_c = |\mathcal{B}_c|$ forms the \emph{presence reward}
\begin{equation}
    \mathcal{R}_{\text{presence}}(c) = \mathbb{I}(\hat{N}_c > 0),
\end{equation}
as well as the \emph{count reward}
\begin{equation}
    \mathcal{R}_{\text{count}}(c) = \exp\!\left( - \, \lvert \hat{N}_c - N_c^* \rvert \right),
\end{equation}
in which $N_c^*$ is the target count given in $c$.  

Beyond category and quantity, object attributes are also verified.  
The \emph{color reward} is defined as  
\begin{equation}
    \mathcal{R}_{\text{color}}(c) = \mathrm{sim}_{\mathrm{color}}(C_{\text{det}}, C^*),
\end{equation}
where $C_{\text{det}}$ is obtained via a CLIP-based~\cite{clip} classifier that evaluates cropped object regions against prompt templates combining each candidate color with the object class name. The top-scoring color is then compared to the target $C^*$ using $\mathrm{sim}_{\mathrm{color}}(\cdot)$.
Similarly, the \emph{orientation reward} assesses angular consistency as  
\begin{equation}
    \mathcal{R}_{\text{ori}}(c) = \mathbb{I}(|\theta_{\text{det}} - \theta^*| \leq \delta_{\theta}),
\end{equation}
where the detected orientation $\theta_{\text{det}}$ is obtained from an orientation-sensitive model~\cite{wang2024orientanythinglearningrobust}, $\theta^*$ is the target in $c$, and $\delta_{\theta}$ specifies the tolerance.  

For 3D spatial reasoning, the \emph{depth reward} evaluates whether the relative depth ordering matches the target:  
\begin{equation}
    \mathcal{R}_{\text{depth}}(c) = \exp\!\left( - \, \lvert d_{\mathrm{rank}} - d_{\mathrm{rank}}^* \rvert \right),
\end{equation}
where $d_{\mathrm{rank}}$ is the rank order inferred via monocular depth estimation~\cite{yang2024depth}, $d_{\mathrm{rank}}^*$ is the target ordering.  

This formulation enforces comprehensive verification of every inclusion constraint $c \in \mathcal{C}_{\text{inc}}$ against the generated image, capturing both appearance fidelity and fine-grained spatial consistency.

\paragraph{Text Content and Localization Reward.}
For prompts that require rendering specific textual content within objects, the reward must jointly assess semantic correctness and spatial placement.  
Given a target object $B_{\text{obj}}$ and required text $T^*$, a global OCR model $\mathcal{F}_{\text{ocr}}$~\cite{cui2025paddleocr30technicalreport,wei2024generalocr} extracts a set of detected text–box pairs $\mathcal{T}_{\text{rec}} = \{(T'_j, B'_j)\}_{j=1}^m$, where $B'_j$ and $B_{\text{obj}}$ denote bounding boxes for detected text regions and the target object, respectively.  
The textual reward is defined by identifying the text instance that best matches $T^*$ in content and is correctly localized within $B_{\text{obj}}$:
\begin{equation}
    \mathcal{R}_{\text{text}}(T^*, B_{\text{obj}}) = \max_{(T'_j, B'_j) \in \mathcal{T}_{\text{rec}}} \left[ \text{sim}(T^*, T'_j) \cdot \text{IoA}(B'_j, B_{\text{obj}}) \right],
    \label{eq:text_reward_joint}
\end{equation}
where $\text{sim}(\cdot)$ measures normalized lexical similarity and $\text{IoA}$ quantifies the degree of spatial containment:
\begin{equation}
    \text{IoA}(B_{\text{text}}, B_{\text{obj}}) = \frac{\text{Area}(B_{\text{text}} \cap B_{\text{obj}})}{\text{Area}(B_{\text{text}})},
    \label{eq:ioa_metric}
\end{equation}
High \emph{text reward} is given only when generated text matches the target string and appears within the correct object bounding box, ensuring prompt fidelity in tasks involving embedded text.

\subsection{Spatial Chain-of-Thought Reasoning}
\label{sec:cot_reasoning}

Although the fine-grained object and text rewards provide reliable verification for individual attributes, determining complex spatial relations between multiple entities requires higher-level reasoning~\cite{ma2023crepe,Benchmarking-spatial,cot-survey}. Simple rule-based geometric checks often struggle with nuanced semantics (e.g., distinguishing ``on'' from ``above'') and with context-dependent layouts that cannot be resolved from geometry alone. To address this, we adopt Qwen2.5-VL~\cite{qwen-vl} as our Chain-of-Thought  reasoning backbone, using verified detection-based signals as grounding to reduce hallucination.

For the spatial relation  between entities $e_A$ and $e_B$, we construct a CoT prompt $P_{\mathrm{CoT}}$ comprising: (1) the target relation $r$, (2) their detected bounding boxes $B_A$ and $B_B$, and (3) the set of attribute rewards for each object from previous stages
\(\{\mathcal{R}_{\mathrm{pres}}, \ldots, \mathcal{R}_{\mathrm{ori}}, \mathcal{R}_{\mathrm{depth}}, \mathcal{R}_{\mathrm{text}}\}\).
By explicitly providing these verifiable signals, the VLM is guided to reason step-by-step: first interpreting each attribute reward relative to the bounding boxes, then performing geometric analysis, and finally inferring whether the relation $r$ holds.

The output of $\mathcal{F}_{\mathrm{vlm}}$ is restricted to a structured format containing the reasoning trace and a final score. We parse the score using a function $\mathcal{P}_{\mathrm{score}}$ to yield the spatial-consistency reward:
\begin{equation}
    \mathcal{R}_{\mathrm{spatial}} = 
    \mathcal{P}_{\mathrm{score}}\big(\mathcal{F}_{\mathrm{vlm}}(P_{\mathrm{CoT}}(r, B_A, B_B, \text{attributes}))\big),
    \label{eq:spatial_reward}
\end{equation}
where $\text{attributes}$ denotes the verified property scores (e.g., presence, orientation, depth, text) for both entities from earlier detection stages.

% To enhance robustness and avoid overfitting to positive cases, we incorporate explicit penalties for satisfied exclusion constraints. For each spatial constraint $c \in \mathcal{C}$, the CoT module produces $\mathcal{R}_{\mathrm{spatial}}(c)$, and the aggregated spatial score is computed as:

To enhance robustness and avoid overfitting to positive cases, we incorporate explicit penalties for satisfied exclusion constraints. For each inclusion constraint $c \in \mathcal{C}_{\mathrm{inc}}$ and exclusion constraint $c \in \mathcal{C}_{\mathrm{exc}}$, both derived from the earlier Prompt Decomposer, the CoT module produces $\mathcal{R}_{\mathrm{spatial}}(c)$, and calculates the aggregated spatial score:

\begin{equation}
    \mathcal{R}_{\mathrm{total}} \;=\; 
    \sum_{c \in \mathcal{C}_{\mathrm{inc}}} \mathcal{R}_{\mathrm{spatial}}^+(c) 
    \;-\; 
    \sum_{c \in \mathcal{C}_{\mathrm{exc}}} \mathcal{R}_{\mathrm{spatial}}^-(c),
\end{equation}
This formulation rewards satisfaction of required relations while penalizing the presence of undesired ones, producing a spatial score that is both semantically informed and grounded in verifiable evidence.

\section{SpatRelBench}
\label{sec:benchmark}

{\renewcommand{\arraystretch}{1.3}
\setlength{\tabcolsep}{3pt} 
\begin{table*}[t!]
\centering
\small 
\caption{\textbf{Quantitative comparison of T2I generation models aligned with different reward models.} 
Results are on GenEval(80-Obj) and SpatRelBench(1k-Obj), where parentheses indicate the number of object categories.
S-Obj: Single object, T-Obj: Two objects, Cnt: Counting, Pos: Positions, Attr-C: Attribute (Color), 
P-Text: Position-Text OCR, C-Text: Counting-Text OCR, Cpx: Complex spatial relations, Ori: Orientation, 3DRel: 3D spatial relations, Overall: average score over all metrics in each dataset. 
\textbf{Bold} denotes the best score, and \underline{underline} denotes the second best.}

\label{tab:main_results}
\begin{tabular}{lccccccccccccc}
\toprule
\multirow{2}{*}{\textbf{Reward Model}} & \multicolumn{7}{c}{\textbf{GenEval (80-Obj)}} & \multicolumn{6}{c}{\textbf{SpatRelBench (1k-Obj)}} \\
\cmidrule(lr){2-8} \cmidrule(lr){9-14}
& S-Obj. & T-Obj. & Cnt. & Color & Pos. & Attr-C. & \textbf{Overall} & P-Text. & C-Text. & Cpx. & Ori. & 3DRel. & \textbf{Overall} \\
\midrule
\rowcolor{gray!20}
\multicolumn{14}{l}{\textit{\textbf{Proprietary T2I models}}} \\

\textbf{GPT Image 1}   & 0.99 & 0.92 & 0.85 & 0.92 & 0.75 & 0.61 & 0.84 & 0.50 & 0.22 & 0.53 & 0.15 & 0.45 & 0.37 \\
\textbf{Seedream 3.0}   & 0.99 & 0.96 & 0.91 & 0.93 & 0.47 & 0.80 & 0.84 & 0.25 & 0.20 & 0.11 & 0.07 & 0.41 & 0.21 \\
\textbf{Qwen-Image}            & 0.99 & 0.92 & 0.90 & 0.88 & 0.76 & 0.77 & 0.87 & 0.22 & 0.21 & 0.32 & 0.07 & 0.32 & 0.23 \\

\midrule
\rowcolor{gray!20}
\multicolumn{14}{l}{\textit{\textbf{Based on Stable Diffusion 3.5}}} \\

\textbf{SD3.5-M}        
& 0.98 & 0.79 & 0.59 & 0.80 & 0.28 & 0.58 & 0.67 
& 0.40 & 0.13 & 0.22 & 0.07 & 0.36 & 0.23 \\

\textbf{+ TextOCR}         
& 0.99 & 0.87 & 0.59 & 0.81 & 0.33 & 0.58 & 0.70  
& \underline{0.48} & 0.24 & 0.23 & 0.09 & 0.38 & 0.28 \\

\textbf{+ PickScore}    
& 0.98 & 0.94 & 0.77 & 0.83 & 0.31 & 0.62 & 0.74 
& 0.36 & 0.15 & 0.28 & 0.06 & 0.33 & 0.24 \\

\textbf{+ QwenVL}       
& 0.99 & 0.58 & 0.48 & 0.81 & 0.24 & 0.55 & 0.61 
& 0.32 & 0.21 & 0.30 & \underline{0.12} & 0.37 & 0.26 \\

\textbf{+ ImageReward}  
& 0.95 & 0.92 & 0.52 & 0.92 & 0.70 & 0.77 & 0.80 
& 0.42 & 0.23 & 0.32 & 0.11 & \underline{0.42} & 0.30 \\

\textbf{+ UnifiedReward}
& \textbf{1.00} & \underline{0.98} & \textbf{0.90} & \underline{0.86} & \underline{0.81} & \underline{0.78} & \underline{0.89} 
& 0.46 & \underline{0.26} & \underline{0.40} & \underline{0.12} & 0.40 & \underline{0.33} \\

\rowcolor{gray!10}
\textbf{+ SpatialReward} 
& \textbf{1.00} & \textbf{0.99} & \underline{0.86} & \textbf{0.96} & \textbf{0.98} & \textbf{0.91} & \textbf{0.95} 
& \textbf{0.51} & \textbf{0.33} & \textbf{0.43} & \textbf{0.26} & \textbf{0.55} & \textbf{0.42} \\

\rowcolor{green!10}
\textbf{Imp. over Baseline} 
& +0.02 & +0.20 & +0.27 & +0.16 & +0.70 & +0.33 & +0.28 
& +0.11 & +0.20 & +0.21 & +0.19 & +0.19 & +0.19 \\

\midrule
\rowcolor{gray!20}
\multicolumn{14}{l}{\textit{\textbf{Based on FLUX}}} \\
\textbf{FLUX1-dev}     & 0.97 & 0.88 & 0.48 & 0.82 & 0.67 & 0.76 & 0.76 & 0.49 & 0.25 & 0.23 & 0.04 & 0.38 & 0.28 \\
\rowcolor{gray!10}
\textbf{+ SpatialReward} & \textbf{1.00} & \textbf{0.99} & \textbf{0.89} & \textbf{0.98} & \textbf{0.99} & \textbf{0.94} & \textbf{0.97} & \textbf{0.63} & \textbf{0.40} & \textbf{0.52} & \textbf{0.32} & \textbf{0.45} & \textbf{0.46} \\
\rowcolor{green!10}
\textbf{Imp. over Baseline} & +0.03 & +0.11 & +0.41 & +0.16 & +0.32 & +0.18 & +0.21 & +0.14 & +0.15 & +0.29 & +0.28 & +0.17 & +0.18 \\
\bottomrule
\end{tabular}
\end{table*}
}

To enable more fine-grained and comprehensive evaluation of spatial relationships in generated images, we introduce \textbf{SpatRelBench}, a benchmark specifically designed to assess spatial consistency in complex scenarios. The evaluation protocol covers five primary dimensions: 
(1) \emph{Complex spatial relations}, which assess the relative positioning and arrangement of multiple objects within a scene; 
(2) \emph{Object orientation}, which checks whether each object is depicted facing the correct direction; 
(3) \emph{Three-dimensional relations}, which evaluate depth ordering and 3D layout consistency among objects; 
(4) \emph{Text-position accuracy}, which verifies whether rendered text appears at the correct location relative to the associated object, using optical character recognition; and 
(5) \emph{Text-counting consistency}, which determines whether the quantity of rendered text across multiple objects matches the prompt specification. 

As illustrated in Fig.~\ref{fig:benchmark}, we extend the evaluation category set beyond the COCO-80~\cite{lin2014coco} classes to include ImageNet-1k~\cite{deng2009imagenet} and Objects365~\cite{objects365} categories, covering both common objects and fine-grained subcategories. Prompts are generated using Gemini-2.5-Pro~\cite{huang2025gemini} and then manually validated to ensure both diversity and correctness. During evaluation, domain-specific expert models are employed to score each dimension. For every sub-task, a binary decision (\emph{correct} or \emph{incorrect}) is recorded, and the overall accuracy is computed by normalizing the number of correct judgments over the total number of requirements. The current release contains approximately 2,000 annotated entries. Built on a modular pipeline, SpatialRelBench is designed to facilitate future expansion to additional categories, tasks, or spatial dimensions, thus providing a challenging and discriminative benchmark for assessing complex spatial consistency in T2I models.

\begin{figure*}[t]
  \centering
  % NOTE: Remember to replace 'placeholder_figure.pdf' with your actual figure file.
  \includegraphics[width=\linewidth]{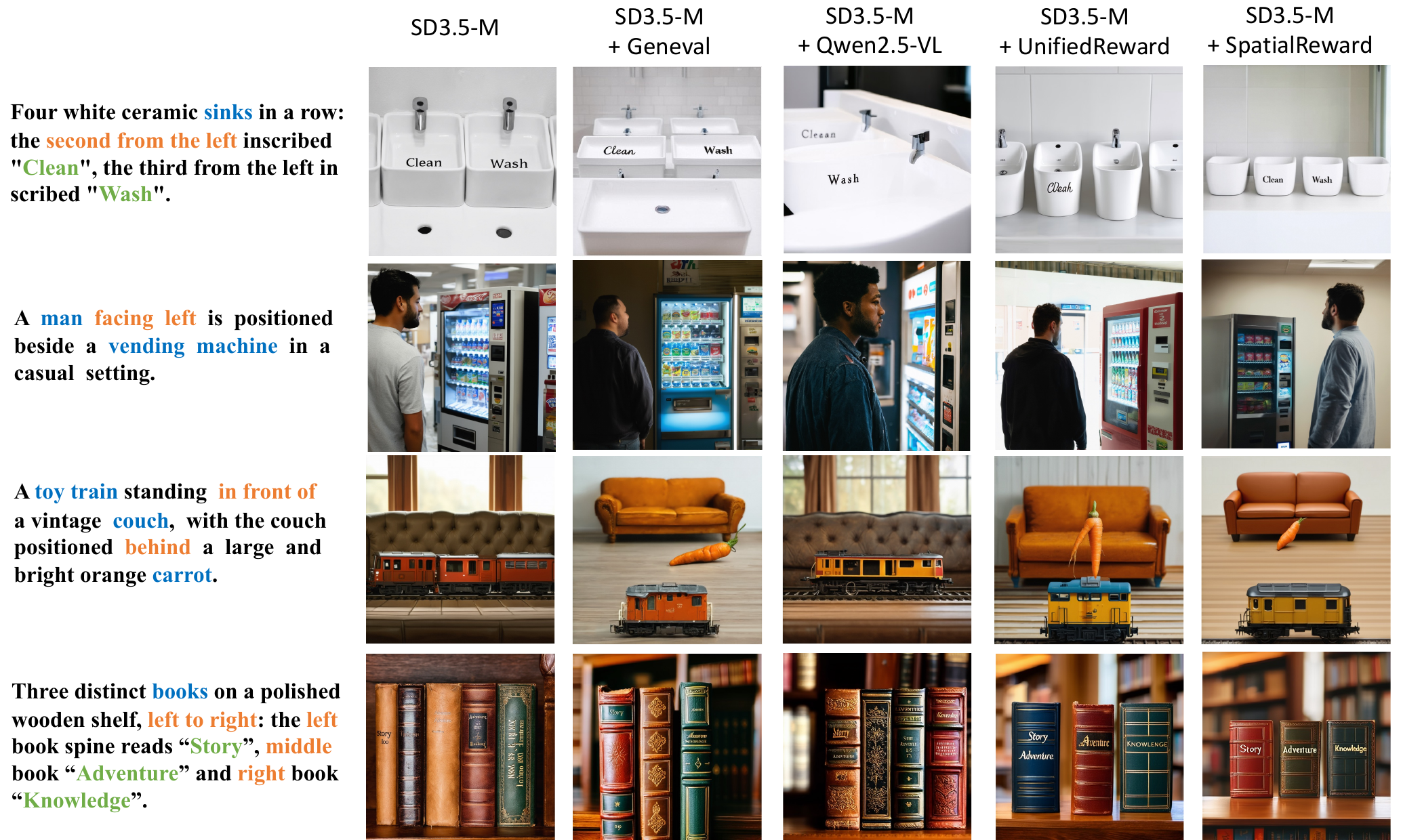} 
  \caption{Qualitative comparison of generated image quality across different methods.}
  \label{fig:paradigm_comparison}
  % \label{fig:framework}
\end{figure*}

\section{Experiments}
\label{sec:experiments}

\subsection{Implementation Details}
\paragraph{Training Configuration.}
We apply reinforcement learning to SD3.5-M~\cite{sd3.5} and FLUX1-dev~\cite{flux}, adopting the GRPO approach provided by the Flow-GRPO~\cite{flowgrpo} framework. During training, we employ a sampling timestep of $T=10$, a group size of $G=24$, a noise level of $a=0.7$, and a fixed image resolution of $512 \times 512$. For evaluation, we increase the timestep to $T=40$. Parameter-efficient tuning is enabled by LoRA with a rank $r=32$ and a scaling factor $\alpha=64$. The KL regularization coefficient $\beta$ is set to 0.04. All models are trained on 16 NVIDIA L20 GPUs.

\paragraph{Evaluation Models and Baselines.}
We evaluate the proposed SpatialReward model by comparing it against a set of established reward models under identical experimental conditions. The baseline set comprises TextOCR~\cite{cui2025paddleocr30technicalreport}, PickScore~\cite{pickscore}, Qwen2.5-VL~\cite{qwen-vl}, ImageReward~\cite{imagereward}, and UnifiedReward~\cite{unifiedreward}. 
To ensure fairness, all reward models are trained on the same dataset, consisting of 100k spatial‑relation prompts automatically generated by GPT‑4o and verified for correctness and diversity. The backbones are optimized via Flow-GRPO framework with identical hyperparameter settings.

\subsection{Quantitative Comparison}

\paragraph{Evaluation on Spatial-Consistency Benchmarks}
As shown in Table~\ref{tab:main_results}, integrating the proposed SpatialReward into both Stable Diffusion and FLUX models yields consistent and substantial improvements across all evaluation dimensions in the standard GenEval benchmark and  SpatialRelBench. On GenEval, SpatialReward enhances performance not only on common object-level metrics but also on complex compositional tasks, while on SpatialRelBench the gains are particularly notable in the challenging dimensions that require fine-grained spatial reasoning, such as multi-object relation understanding, orientation accuracy, depth ordering, and text–position alignment. The improvements manifest across both benchmarks and for different backbone models, demonstrating that SpatialReward effectively enforces spatial coherence and semantic fidelity in generated images. These results confirm the robustness and generalizability of our approach, and highlight the discriminative power of SpatialRelBench in revealing performance gaps that remain hidden under single-dimensional evaluation protocols.

{\renewcommand{\arraystretch}{1.4}
\begin{table}[t]
\centering
\small
\caption{General-Purpose comparison on Wise, DPG, Aesthetic, and PickScore metrics.}
\label{tab:wise_dpg}
\begin{tabular}{lcccc}
\toprule
\textbf{Model} & \textbf{Wise} & \textbf{DPG} & \textbf{Aesthetic} & \textbf{PickScore} \\
\midrule
\textbf{SD3.5-M}       & 0.45 & 83.96 & \textbf{5.39} & 22.34 \\
\textbf{+ SpatialReward}     & \textbf{0.46} & \textbf{84.08} & 5.23 & \textbf{22.52} \\
\midrule
\textbf{Flux1-dev}           & 0.50 & 83.84   & 6.13 & 22.45 \\
\textbf{+ SpatialReward}     & \textbf{0.52} & \textbf{84.19}   & \textbf{6.15} & \textbf{23.22} \\
\bottomrule
\end{tabular}
\end{table}
}

{\renewcommand{\arraystretch}{1.25}
\begin{table}[t]
\centering
\small
% \caption{Correlation and classification performance between reward model scores and human judgments of spatial consistency on 500 image–prompt pairs. Higher values indicate better alignment with human evaluation. Accuracy computed at threshold $\tau=0.5$.}
\caption{Correlation and accuracy with human spatial-consistency judgments; accuracy measured at threshold $\tau=0.8$.}
\label{tab:human_alignment}
\begin{tabular}{lccc}
\toprule
\textbf{Reward Model} & \textbf{Spearman $\rho$} & \textbf{Pearson $r$} & \textbf{Accuracy} \\
\midrule
\textbf{CLIPScore} & 0.42 & 0.40 & 0.68 \\
\textbf{ImageReward} & 0.48 & 0.45 & 0.70 \\
\textbf{UnifiedReward} & 0.51 & 0.49 & 0.72 \\
\textbf{VisionReward} & 0.55 & 0.53 & 0.74 \\
% \rowcolor{gray!10}
\textbf{SpatialReward}  & \textbf{0.63} & \textbf{0.61} & \textbf{0.79} \\

\bottomrule
\end{tabular}
\end{table}
}

\paragraph{Evaluation on General-Purpose Metrics}
To assess the generalization of SpatialReward beyond spatial-focused benchmarks, we evaluated it using widely adopted metrics: overall fidelity (Wise~\cite{niu2025wise}, DPG~\cite{hu2024ella-DPG}), visual appeal (Aesthetic~\cite{aesthetics}), and prompt–image alignment (PickScore~\cite{pickscore}). With both Stable Diffusion and FLUX backbones, SpatialReward matched or exceeded baseline performance across all metrics, except for a slight drop in Aesthetic. This suggests reward optimization maintains overall visual quality and semantic alignment. Gains on Wise and PickScore reflect improved holistic prompt–image alignment, while stable Aesthetic and DPG scores indicate enhanced spatial reasoning without perceptual loss.

\subsection{Human Alignment on Spatial Consistency}
We evaluated whether SpatialReward aligns more closely with human perception of spatial consistency through a study on 500 prompt–image pairs from SpatialRelBench, each generated by GPT-4o and labeled by annotators as correct or incorrect in depicting the intended spatial relations. Compared with representative general-purpose reward models, SpatialReward showed the highest correlation with human judgments (Spearman’s $\rho$~\cite{spearman1904}, Pearson’s $r$~\cite{pearson1895}) and the best classification accuracy at a fixed threshold $\tau=0.8$, confirming that explicitly modeling fine-grained spatial relations produces assessments more consistent with human perception and serves as a reliable metric for complex spatial tasks in text-to-image generation.

% {\renewcommand{\arraystretch}{1.25}
% \begin{table}[t]
% \centering
% \caption{Ablation study on SpatialReward components. Scores are accuracy (\%) for final T2I models trained with different reward model variants across three benchmarks.}
% \label{tab:ablation_results_redesign}
% \resizebox{\linewidth}{!}{%
% \begin{tabular}{lccc}
% \toprule
% \textbf{Removed Component} &  \textbf{GenEval} & \textbf{SpatRelBench} & \textbf{T2I-CBench} \\
% \midrule
% \rowcolor{gray!15}
% Full SpatialReward         & \textbf{85.4} & \textbf{84.1} & \textbf{84.8} \\
% – Constraint Parsing       & 81.6 & 80.3 & 80.7 \\
% – Attribute Verification   & 80.9 & 79.6 & 80.4 \\
% – Spatial Reasoning Step   & 79.5 & 78.1 & 78.8 \\
% – Exclusion Handling       & 80.2 & 79.1 & 79.7 \\
% \bottomrule
% \end{tabular}%
% }
% \end{table}
% }

{\renewcommand{\arraystretch}{1.4}
\begin{table}[t]
\centering
% \caption{Impact of removing individual components from the SpatialReward framework. 
% Scores (accuracy, \%) are reported on three benchmarks. 
% Higher values indicate better alignment between generated images and target spatial relations.}
% \caption{Impact of removing components from the SpatialReward. 
% Accuracy are reported on three benchmarks: GenEval, 
% SpatRel. : (\emph{SpatialRelBench}), and T2IComp. : (\emph{T2I-CompBench}). 
% For T2I-CompBench, results correspond to the mean accuracy over its 2D and 3D spatial-consistency tasks.}
\caption{Ablation results for SpatialReward. Scores (accuracy) are reported for GenEval, SpatRel: SpatialRelBench, and T2IComp: T2I-CompBench. T2I-CompBench values are averaged over its 2D and 3D spatial-consistency tasks.}

\label{tab:ablation_results_key}
\resizebox{\linewidth}{!}{%
\begin{tabular}{lccc}
\toprule
% \textbf{Removed Component} & \textbf{GenEval} & \textbf{SpatRelBench} & \textbf{T2I-CBench} \\
\textbf{Removed Component} & \textbf{GenEval} & \textbf{SpatRel.} & \textbf{T2IComp.} \\
\midrule
% \rowcolor{gray!15}
\textbf{Full SpatialReward}                & \textbf{95.2} & \textbf{37.1} & \textbf{50.1} \\
– Exclusion Constraints           & 90.5 & 25.9 & 45.9 \\
– Expert Detection                 & 70.3 & 21.6 & 39.2 \\
– CoT Reasoning                    & 94.2 & 27.9 & 47.5 \\
\bottomrule
\end{tabular}%
}
\end{table}
}

\subsection{Ablation Study}

To assess the contribution of each component within SpatialReward, we conduct ablation experiments by selectively removing three key modules: \emph{Exclusion Constraints}, \emph{Expert Detection}, and \emph{Chain-of-Thought Reasoning}. The resulting accuracy variations across GenEval, SpatialRelBench, and T2I-CompBench are summarized in Table~\ref{tab:ablation_results_key}. 

\paragraph{Effect of Exclusion Constraints}
Removing the exclusion constraint, which introduces negative samples to penalize undesired spatial configurations, consistently reduces accuracy across all benchmarks. Incorporating such penalties improves reward robustness, mitigates over-optimization, and prevents reward hacking that could degrade real-world performance. This mechanism is particularly beneficial for prompts containing distractor objects, preventing performance degradation in realistic settings.

\paragraph{Effect of Expert Detection}
Omitting the expert detection stage, which performs fine-grained verification of object presence, attributes, and rendered text, results in the largest accuracy decline across all benchmarks. This outcome substantiates our core principle that spatial reasoning benefits significantly from verified reward signals, as domain-specific detectors provide reliable and interpretable evidence that general-purpose vision–language models alone cannot consistently ensure.

\begin{figure}[!t]
  \centering
  \includegraphics[width=\columnwidth]{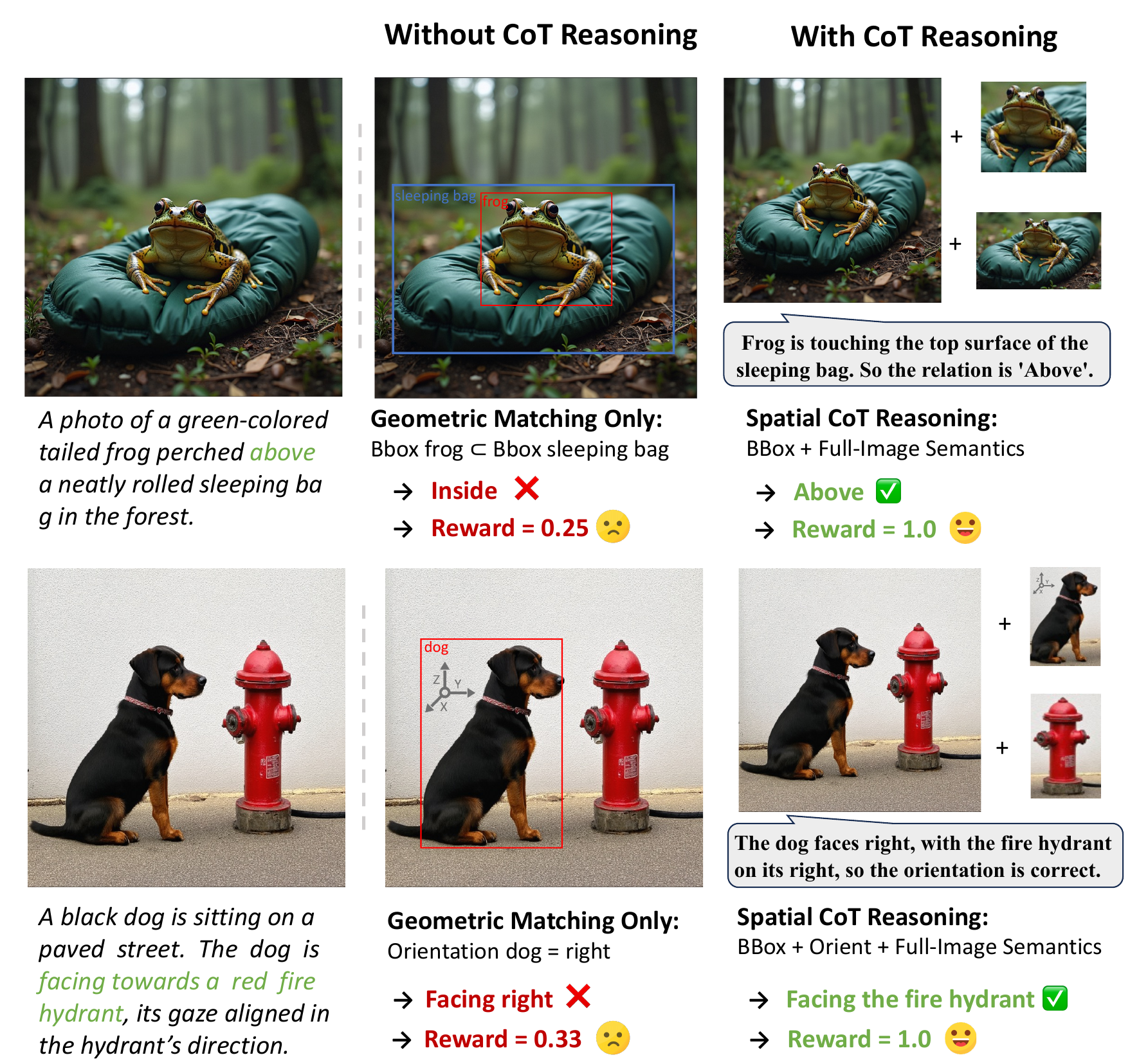}
\caption{Effect of CoT reasoning in spatial relations. CoT combines bounding boxes, orientation, and scene semantics, yielding correct classifications where detected-only matching fails.}
  \label{fig:cot_cases}
  \vspace{-1mm} % 缩小图与正文之间的竖直间距
\end{figure}

\paragraph{Effect of Chain-of-Thought Reasoning}
Eliminating the Chain-of-Thought reasoning process, which integrates verified object attributes with geometric analysis to infer spatial relations, results in moderate accuracy reductions. Although the quantitative decline is smaller than that observed for other components, qualitative case studies (Fig.~\ref{fig:cot_cases}) highlight CoT’s critical role in complex scenarios. Examples include distinguishing “above” from “inside” and resolving intricate multi-object arrangements that rule-based heuristics cannot effectively handle.

\section{Conclusion}
We presented SpatialReward, a spatial-consistency verifiable reward model, and SpatialRelBench, a benchmark for fine-grained evaluation of spatial relations in text-to-image generation. By combining constraint parsing, expert-based verification, and chain-of-thought reasoning, SpatialReward effectively enforces spatial fidelity in complex scenes. Experiments on both SpatialRelBench and general evaluation metrics, along with human alignment studies, show that our method surpasses existing reward models in capturing spatial correctness without compromising overall image quality. The results support the use of verifiable spatial reward as a reliable and adaptable objective for future text-to-image generation models.

% \input{sec/appendix}
% \section*{Acknowledgements}
% This work was supported in part by National Natural Science Foundation of China under Grant No.624B2128 and No.62222211.
\section*{Acknowledgements}
This work was supported by the National Natural Science Foundation of China (NSFC) under Grant No. U24A20326, and the Joint Fund Project "End-Cloud Collaborative Lightweight Autonomous Intelligence for Content Generation".
% \newpage

{
\small
\renewcommand\UrlFont{\color{Gray}\ttfamily}
\bibliographystyle{ieeenat_fullname}
\bibliography{main}
}

\newpage
\appendix

%%%%%%%%% BODY TEXT - ENTER YOUR RESPONSE BELOW
% \section{Examples of SpatialBench}

% \input{sec/appendix}

% In Fig.~\ref{fig:outdoor} and \ref{fig:indoor}, we show outdoor and indoor examples from the SpatialBench, respectively. Our benchmark constructs misleading hard negative captions to evaluate the model's ability to understand various spatial concept, including depth, orientation, spatial state, object relationships and object size, etc.

% \begin{figure*}[t]
%     \centering
%     \includegraphics[width=0.72\linewidth]{1.png}
%     \includegraphics[width=0.72\linewidth]{2.png}
%     \includegraphics[width=0.72\linewidth]{3.png}
%     % \vspace{-1.5\baselineskip}
%     \caption{Examples from the outdoor subset of SpatialBench.}
%     \label{fig:outdoor}
% \end{figure*}

% \begin{figure*}[t]
%     \centering
%     \includegraphics[width=0.72\linewidth]{11.png}
%     \includegraphics[width=0.72\linewidth]{22.png}
%     \includegraphics[width=0.72\linewidth]{33.png}
%     % \vspace{-1.5\baselineskip}
%     \caption{Examples from the indoor subset of SpatialBench.}
%     \label{fig:indoor}
% \end{figure*}

\end{document}